\documentclass[letterpaper, 10 pt, conference]{ieeeconf}  

\usepackage{graphicx}
\usepackage{lipsum}

\usepackage{xcolor}
\usepackage{hyperref}

\IEEEoverridecommandlockouts                              

\title{\LARGE \bf
Autonomous Control of a Tendon-driven Robotic Limb with Elastic Elements Reveals that Added Elasticity can Enhance Learning
}

\author{Ali Marjaninejad$^{1}$, Jie Tan$^{2}$, and Francisco Valero-Cuevas$^{3}$, \it{Senior Member, IEEE}
\thanks{$^{1}$A. Marjaninejad is with University of Southern California, Los Angeles, Ca 90089 USA
        {\tt\small e-mail: marjanin@usc.edu}}%
\thanks{$^{2}$J. Tan is with Google Brain, Mountain View, CA, 94043.
        {\tt\small e-mail: jietan@google.com}}%
\thanks{$^{3}$F. Valero-Cuevas is with University of Southern California, Los Angeles, Ca 90089 USA
        {\tt\small (corresponding author) email-: valero@usc.edu; phone: 213-740-4219}}%
}

\begin{document}

\maketitle
\thispagestyle{empty}
\pagestyle{empty}

\begin{abstract}

Passive elastic elements can contribute to stability, energetic efficiency, and impact absorption in both biological and robotic systems.
They also add dynamical complexity which makes them more challenging to model and control.
The impact of this added complexity to autonomous learning has not been thoroughly explored. This is especially relevant to tendon-driven limbs whose cables and tendons are inevitably elastic.
Here, we explored the efficacy of autonomous learning and control on a simulated bio-plausible tendon-driven leg across different tendon stiffness values.
We demonstrate that increasing stiffness of the simulated muscles can require more iterations for the inverse map to converge but can then perform more accurately, especially in discrete tasks.
Moreover, the system is robust to subsequent changes in muscle stiffnesses and can adapt on-the-go within 5 attempts.
Lastly, we test the system for the functional task of locomotion, and found similar effects of muscle stiffness to learning and performance.
Given that a range of stiffness values led to improved learning and maximized performance, we conclude the robot bodies and autonomous controllers---at least for tendon-driven systems---can be co-developed to take advantage of elastic elements. 
Importantly, this opens also the door to development efforts that recapitulate the beneficial aspects of the co-evolution of brains and bodies in vertebrates.
\end{abstract}

\section{INTRODUCTION}

Elastic elements are known to contribute in a passive way to a number of advantageous mechanical properties of robotic and biological systems. These include absorbing impacts, storing energy and postural stability. By absorbing impacts, elastic elements reduce noise and prevent damage to the structural elements and actuators (linkages, hinges and motors in robots; and bones, joints, and musculotendons in animals) or the environment~\cite{ananthanarayanan2012towards, wensing2017proprioceptive, mazumdar2016parallel, jain2011controlling}. Also, opposing pairs of elastic elements act like proportional controllers (that can only pull) that can passively grant postural stability~\cite{pratt2002low,zhou2012survey,babikian2016slow,milner2002contribution,mussa1985neural,osu1999multijoint,perreault2002voluntary,perreault2001effects}. It is also known that great energetic efficiency can be achieved by storing and timely release of energy in elastic elements~\cite{mazumdar2016parallel,seok2014design,hurst2008role}.

These benefits, however, come at a cost. They can add nonlinearities, hysteresis and oscillatory modes to the system dynamics and, in general, make it harder to model and find accurate and robust analytical control solutions~\cite{tan2012soft}. This is especially the case for analytical control methods that require precise models of the plant and the environment to operate accurately~\cite{nguyen2012online, doyle1978guaranteed} which is, in general, infeasible for most real-world plants and problems. Moreover, the mechanical properties of elastic materials are more often susceptible to changes in environmental (e.g., temperature), and use-case (e.g., wear and tear) factors.

An alternative approach to the control of plants with elastic elements would be to use control methods that do not depend on prior models, are data-driven, autonomous, or adaptable on the fly.  However, the performance of these methods in dealing with added dynamical complexities introduced with the elastic elements has not been thoroughly explored. Moreover, the robustness of such methods to changes in stiffness values or operation in different functional regimes (e.g., nonlinear springs) needs to be addressed as well.  This is an under-studied problem especially on bio-inspired, tendon-driven systems.

Tendon-driven systems are particularly interesting because they can offer great functional agility and versatility and freedom of design (e.g., actuator placement and tendon routing)~\cite{marjaninejad2019should,marjaninejad2018model,valero2016fundamentals,king2018design,sueda2008musculotendon}. Moreover, they can help us better understand and even approach the diversity and functional versatility of animals by shedding some light on governing principles of vertebrate form and function~\cite{marjaninejad2019autonomous, marjaninejad2018analytical}.

These systems, on the other hand, are harder for engineers to model and analytically control for a number of reasons. To begin with, they are simultaneously under- and over-determined as, respectively, multiple muscle forces can produce a same net torque at a joint, yet a single joint rotation sets the lengths of all muscles that cross it.  Thus, it can be challenging to find solutions that satisfy all the constraints imposed by tendons and by task specifications at the same time~\cite{marjaninejad2019autonomous, valero2016fundamentals}. Moreover, the fact that their actuators are not directly operating on the degrees of freedom (as is the case in joint-driven systems), makes it challenging to use an off the shelf controller (such as a simple PID setup) without having access to dynamical equations of the system or a forward or inverse kinematics model~\cite{marjaninejad2019simple}. Also, these tendon-driven systems often require accurate modeling and control strategies for applications such as animation of life-like figures~\cite{lee2019scalable}, control of anatomical limbs to understand neurological conditions ~\cite{kurse2012extrapolatable,niu2017neuromorphic,jalaleddini2017neuromorphic}, or functional electrical stimulation of limbs (e.g.,~\cite{loeb2006bion} or ~\cite{peckham2005functional}). 

Here, we explored the efficacy of autonomous learning and control on a simulated bio-plausible tendon-driven leg across different tendon stiffness values. For he sake of generality, in this first study, we used two autonomous learning algorithms---one that builds a data-driven explicit kinematics model of the limb vs. one that uses end-to-end learning (see Methods)---to gauge the effect of elasticity of the actuators on learning and performance.
Our results show that autonomous learning (both with an explicit inverse map and end-to-end) could learn to control the limb across all stiffness values. Our results also show that an appropriate value of added stiffness can enhance the learning and precision in all cases and even exhibit emergence of lower energy consumption. This is of great significance because the elasticity that is inherent to some types of plants (i.e., tendon-driven systems) can now be leveraged to improve learning and performance.

\section{Methods}
In this paper, we studied how adding elastic elements affects autonomous learning in a two-joint three-tendons simulated limb (similar to~\cite{marjaninejad2019autonomous,marjaninejad2019simple}) in the MuJoCo environment~\cite{todorov2012mujoco}(Fig.~\ref{fig:Fig1}.a). The muscle model we use consist of a contractile element with Force-Length-Velocity properties~\cite{todorov2012mujoco,valero2016fundamentals}, a small parallel damper (100 Ns/m) and a parallel elastic element with stiffness value `K' (see Fig.~\ref{fig:Fig1}.b). Specifically, we studied the convergence of the inverse kinematics map, how its performance accuracy changes with stiffness,  as well as its adaptability when learning with one stiffness value and then performing using a different value.

As for learning, we used our autonomous few-shot hierarchical learning algorithm General-to-Particular (G2P)~\cite{marjaninejad2019autonomous}, and the end-to-end Proximal Policy Optimization (PPO) autonomous learning algorithm~\cite{schulman2017proximal}. G2P is a hierarchical autonomous learning algorithm that, on its lower-level, creates an inverse kinematics map using output kinematics collected from an initial random set of actuation commands (motor babbling). Systems that use an explicit kinematics model are, in general, easier to study and interpret, more data efficient and can generalize to a wider range of tasks; however, they can suffer from inaccuracies in the model especially during complex dynamical interactions (e.g., contact dynamics, injury to the body, or changes in the environment)~\cite{marjaninejad2019simple,marjaninejad2019autonomous,kwiatkowski2019task,cully2015robots, yang2019data,nguyen2011model}. Systems that perform end-to-end learning (such as PPO), on the other hand, usually require larger number of samples to learn to perform a task, are harder to interpret due to their implicit modeling, and usually cannot generalize well across tasks~\cite{lillicrap2015continuous,schulman2015trust, schulman2017proximal,heess2017emergence,mnih2015human}. These methods, however, can achieve better asymptotic performance even in challenging tasks.

\subsection{Simulated experiments}

\begin{figure}
\centering
\includegraphics[width=.95 \linewidth]{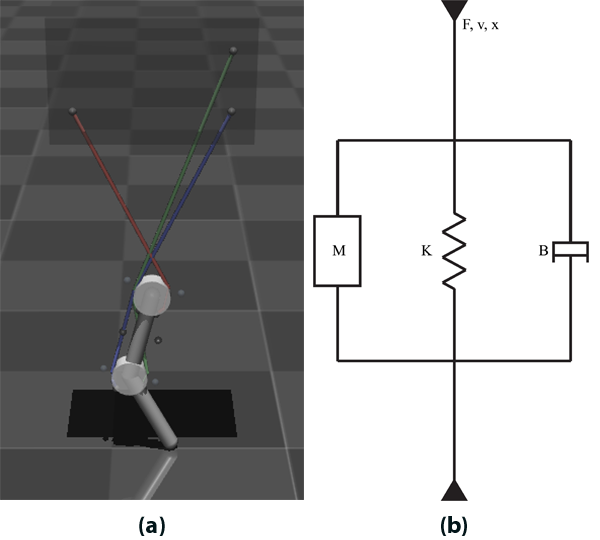}
\caption{(a) The studied tendon-driven limb in MuJoCo environment. (b) each musculotendon consists of a muscle model (M), elastic element (K), and a damper (B).}
\label{fig:Fig1}
\end{figure}

For this study, we have performed three set of simulated experiments. In all simulations, elastic elements are considered as parallel elements with each musclotendon (Fig.~\ref{fig:Fig1}.b); the stiffness value of all elements are equal for each simulation and refereed to as “stiffness”. The details for each of these set of simulations are provided below.

\subsubsection{Controlling the limb with different stiffness values in the muscle model}
In this simulation, for each stiffness value, we first randomly activated muscles and recorded the resulting kinematics (motor babbling~\cite{marjaninejad2019autonomous}) for 3 minutes (100 samples per second). The recorded kinematics are joint angles, angular velocities, and angular accelerations for both joints (a vector of 6 values). Next, we trained a Multi-Layer Perceptron (MLP) Artificial Neural Network (ANN; one hidden layer with 15 neurons; trained for 20 epochs; 80$\%$ training 20$\%$ validation; loss function: MSE, optimizer: ADAM) with kinematics as input and activations as output to form the inverse kinematics map (similar to ~\cite{marjaninejad2019autonomous,marjaninejad2019simple}). Finally, this inverse map was used to control the system to perform two tasks: Cyclical and Point-to-point movements. 

\paragraph{Cyclical movements}
In this task, the system was prescribed to move to generate a perfect circle in its configuration space (joint angle space). I.e., Joint angles change sinusoidal with $\pi/2$ phase difference. The frequency of these cyclical movements was set to 0.7 Hz and the task was continued for 21 cycles (total of 30 seconds).

\paragraph{Point-to-point movements}
Unlike the cyclical movements task, which is a smooth continuous task, the point-to-point task is consisted of discrete joint angle locations connected with rapid movements. In this task, 10 independent random angles (sampled from a uniform distribution within the range of each joint) are selected for each joint. The system then is commanded to go to each joint angle pair and stay there for 3 seconds (total of 30 seconds).
Similar to our previous work~\cite{marjaninejad2019autonomous}, we chose these tasks since they cover both extremities in the movement spectrum between continuous and smooth movements and discrete movements with fast transitions.
For each joint, we calculate the error as the Root Mean Square Error (RMSE) of the difference between the joint angle and the desired angle in Radians. We disregard the error for the first 25$\%$ of the signal to make sure any initial condition effect is washed out~\cite{marjaninejad2019autonomous,marjaninejad2019simple}.

\subsubsection{Adaptability to changes in stiffness}
Stiffness value of an elastic element can change as a function of many physical factors such as temperature, wear and tear, etc. This can potentially endanger performance of the autonomous control of a system even if the system performs accurately in absence of any changes. This task is designed to study this effect as well as studying the feasibility of adaptive learning on-the-go (without a need to stop the system and redo the babbling) to compensate for these changes.

Here, we first perform the motor babbling for a system with an initial stiffness value (let’s call it “A”) and train the inverse map with the collected data. Then, we change the stiffness value (to let’s say “B”) and command the system to perform a cyclical movement attempt (described above). After each attempt, we concatenate all collected data and refine the inverse map using the cumulative data (refinement phase of G2P~\cite{marjaninejad2019autonomous}). Here, we are showing results for up to 5 refinements for a system the stiffness value of which has changed (from “A” to “B”) as well as systems that performed both babbling and refinements with the same stiffness value (“A” to “A” and “B” to “B”) to provide better insight for a better comparison of the adaptation performance.

\subsubsection{Functional task of locomotion}
Studying the ability of our system in creating an inverse kinematics map for different stiffness values provides great insight into understanding the effects of stiffness on control and learning. However, a precise inverse map does not necessarily mean better performance in performing functional tasks that also features contact dynamics~\cite{marjaninejad2019simple}. Also, most autonomous control methods do not use an explicit inverse kinematics map. Therefore, it is important to study the effects of stiffness on the performance of the system for a functional task. We chose a locomotion task that entertains contact dynamics, deals with gravity and inertia, and yields a reward as a measure of success. For this task, the limb is connected to a chassis that can move in x-axis (forward-backward) with friction to stop the system from floating. The system can also move on y-axis (up-down) where it is assisted with a spring-damper mechanism (similar to a gantry~\cite{marjaninejad2019simple}). Please see the Supplementary Video for the task in action.

We have performed this task with two leading algorithms in autonomous learning. First, the G2P algorithm~\cite{marjaninejad2019autonomous,marjaninejad2019simple}, which is specifically designed to handle challenging task of learning and adaptation with no prior model and only using limited experience (which is a need in most real-world applications) and has proved to work well on the tendon-driven systems. Second, we have chosen the PPO algorithm, which is one of the leading end-to-end learning methods: for each observation, predicts activations that will yield high reward. The G2P implementation was in a faithful manner to the original paper~\cite{marjaninejad2019autonomous}, babbling time was selected to be 3 minutes, and the exploration-exploitation reward threshold was set to 3 meters of the chassis movement in the forward direction. Each attempt would be consisted of 10 steps (1.3 seconds each). For PPO, we used the “PPO1” implementation from Open AI’s stable baselines repository. We run the training for 5000 episodes (1000 samples each; sampling rate: 100Hz).

\section{Results}

\subsubsection{Controlling the limb for different stiffness values}
\begin{figure}
\centering
\includegraphics[width=1 \linewidth]{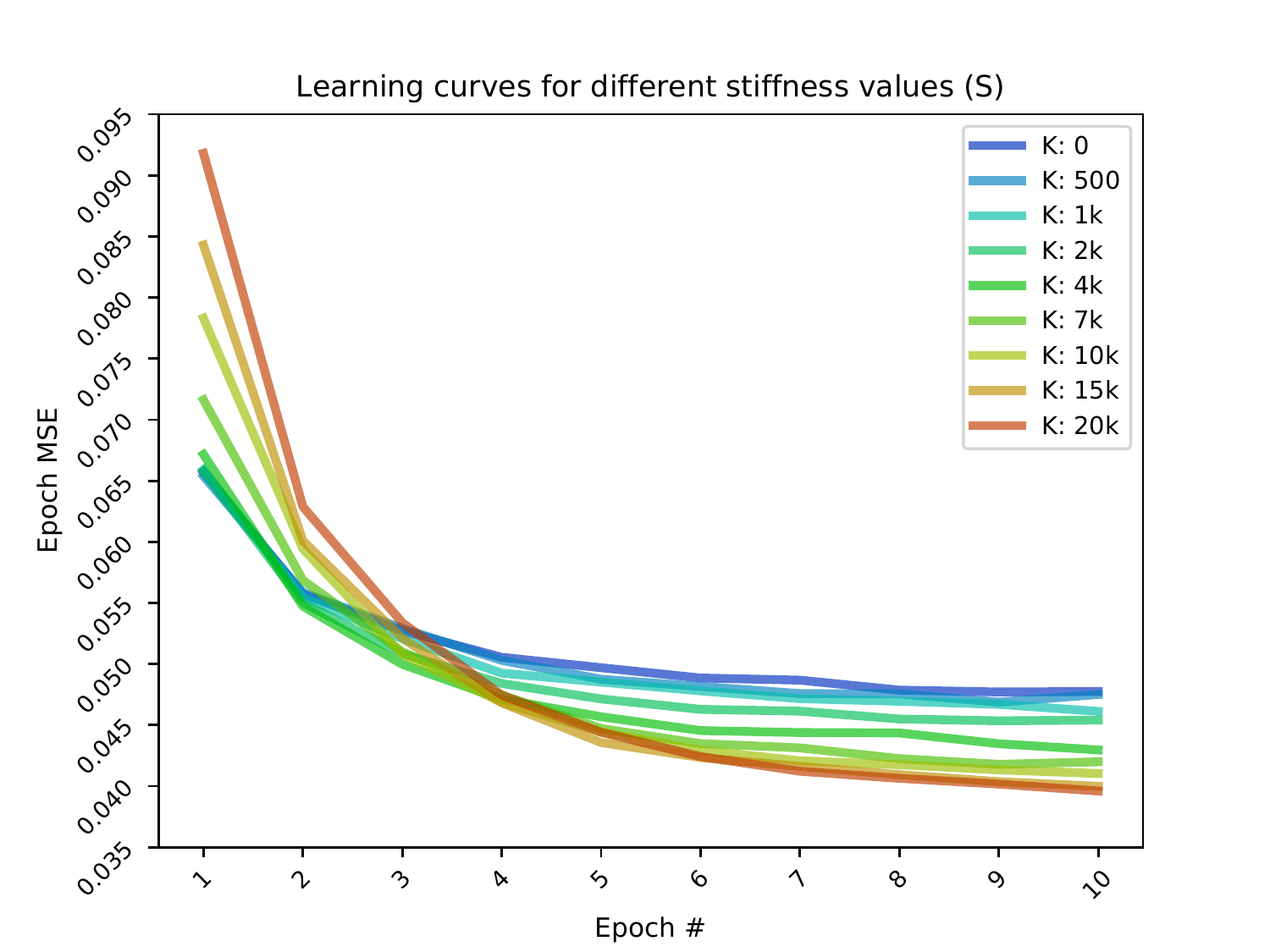}
\caption{MSE over the training data as a function of the epoch number across stiffness values (Average of 50 Monte Carlo runs).}
\label{fig:Fig2}
\end{figure}

Fig.~\ref{fig:Fig2} shows the MSE over the training data as a function of the epoch number across stiffness values. We see a consistent pattern in the training error curves in which systems with higher stiffness values start with larger error, yet once enough training rounds (epochs) are performed, they exhibit the smallest training errors. This pattern can be explained by the fact that more stiffness will add more dynamics to the system which initially makes it harder for the ANN to catch, but once converged, these extra dynamics can reduce the size of the solution space~\cite{cohn2018feasibility,marjaninejad2019should,valero2016fundamentals} (less ambiguity caused by the under-determined nature of the system) and therefore make more precise predictions.
However, these MSE values only show how well the ANN could fit to the training data coming from the motor babbling (see Methods). Therefore, to study its performance across tasks, we now focus on the results collected from the cyclical and point-to-point tasks.
Fig.~\ref{fig:Fig3} shows RMSE values for this simulation across all tested stiffness values (also see the Supplementary Video). We see that stiffness in the range of 2k-10k N/m can significantly improve performance compare to zero stiffness or very high stiffness values. This improvement is even more significant for the point-to-point task which is explained by the fact that this task is more prove to the adverse effects of control in under-determined systems (see Discussion and ~\cite{marjaninejad2019should,valero2016fundamentals}).

\begin{figure*}
\centering
\includegraphics[width=1 \linewidth]{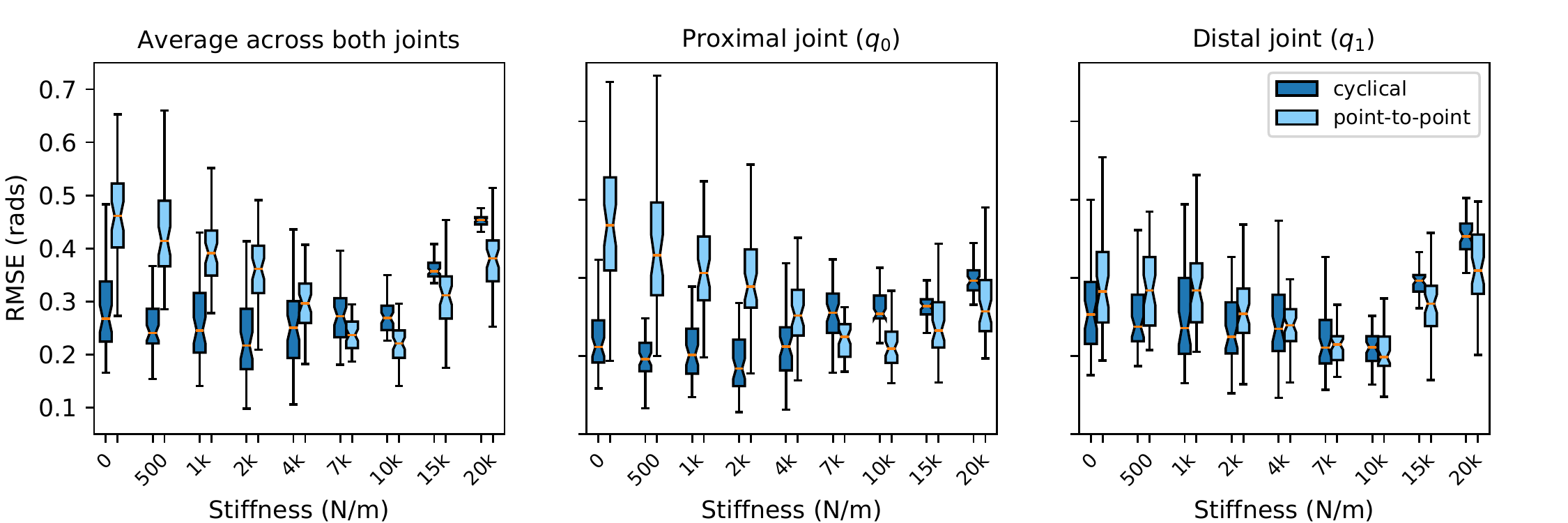}
\caption{RMSE of joint angles as a function of stiffness for cyclical (dark blue) and point-to-point (light blue) tasks. 50 Monte Carlo runs for each case.}
\label{fig:Fig3}
\end{figure*}

\begin{figure*}
\centering
\includegraphics[width=1 \linewidth]{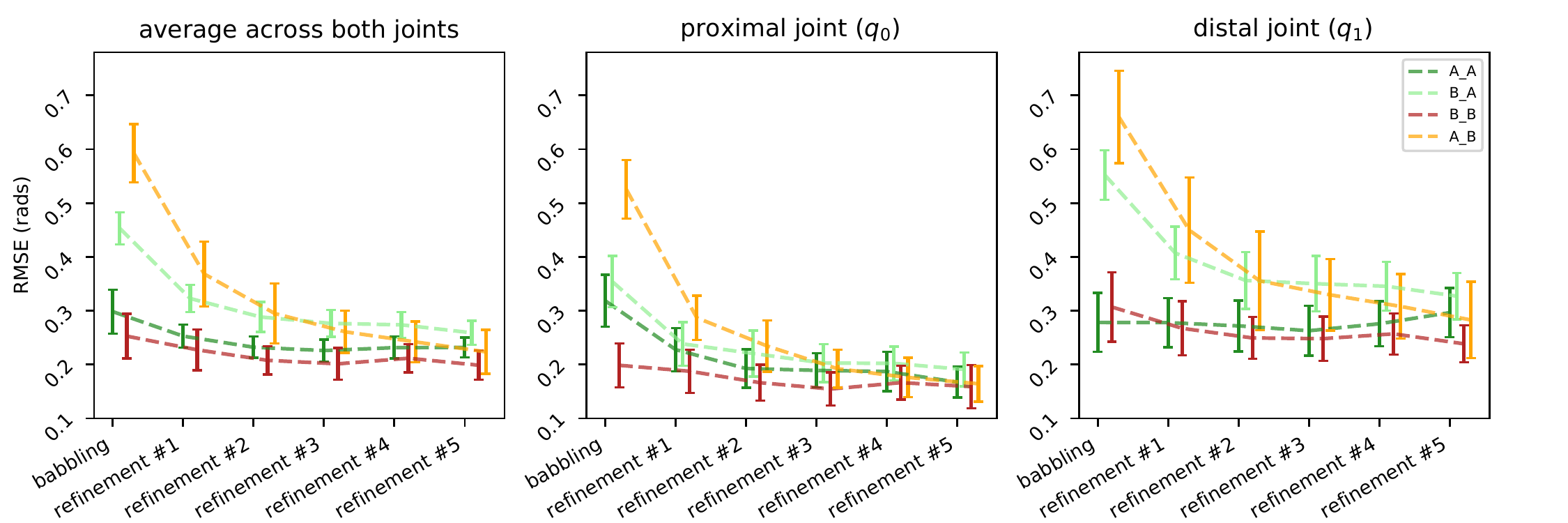}
\caption{RMSE of the systems trained and tested with different stiffness values. A: 7K N/m and B: 2K N/m. (A\_B (orange): trained with A, refined and tested with B; B\_A (light green) the other way around), as well as the performance of systems trained, refined, and tested with the same stiffness values for baseline comparison (A\_A and B\_B, red and dark green, respectively), 50 Monte Carlo runs for each case.}
\label{fig:Fig4}
\end{figure*}
\subsubsection{Adaptability to changes in stiffness}
Fig.~\ref{fig:Fig4} shows the performance of the system trained and tested with different stiffness values as well as its progress through refinements. Fig.~\ref{fig:Fig4} also shows the performance of systems trained, refined, and tested with the same stiffness values for comparison. In Fig.~\ref{fig:Fig4}, A and B correspond to 7K N/m and 2K N/m, respectively. Adaptability between other stiffness values, in general, also followed the same pattern (in all error bars and error shades in this paper, end to end height of whiskers/shades are equal to one standard deviation of the data).
Fig.~\ref{fig:Fig4} results show that it is feasible for a system to create and initial inverse map and then adapt on-the-go while converging to similar performance measures as if it did not have a change. This is important since it will prove the use of elastic elements that are subject to change due to physical features (temperature, wear and tear, etc.) to be feasible in real-world robotic systems. We used G2P here and showed the feasibility of adaptation on-the-go to the changes in the tendon stiffness values. However, we want to underline that other adaptive learning methods can also be used (e.g., ~\cite{kwiatkowski2019task,cully2015robots,bongard2006resilient}).

\subsubsection{Functional task of locomotion}
In this section, we study the results for the functional task of locomotion for two autonomous learning algorithms, namely, G2P and PPO (see methods). It is important to note that the focus of this section is to study the potential effects and contributions of the elastic element in an unbiased manner and not maximizing performance (e.g., using feedback to minimize the error~\cite{marjaninejad2019simple},  finding the optimal solution or the most efficient one) or modifying the algorithms to do so.

\begin{figure*}
\includegraphics[width=1 \linewidth]{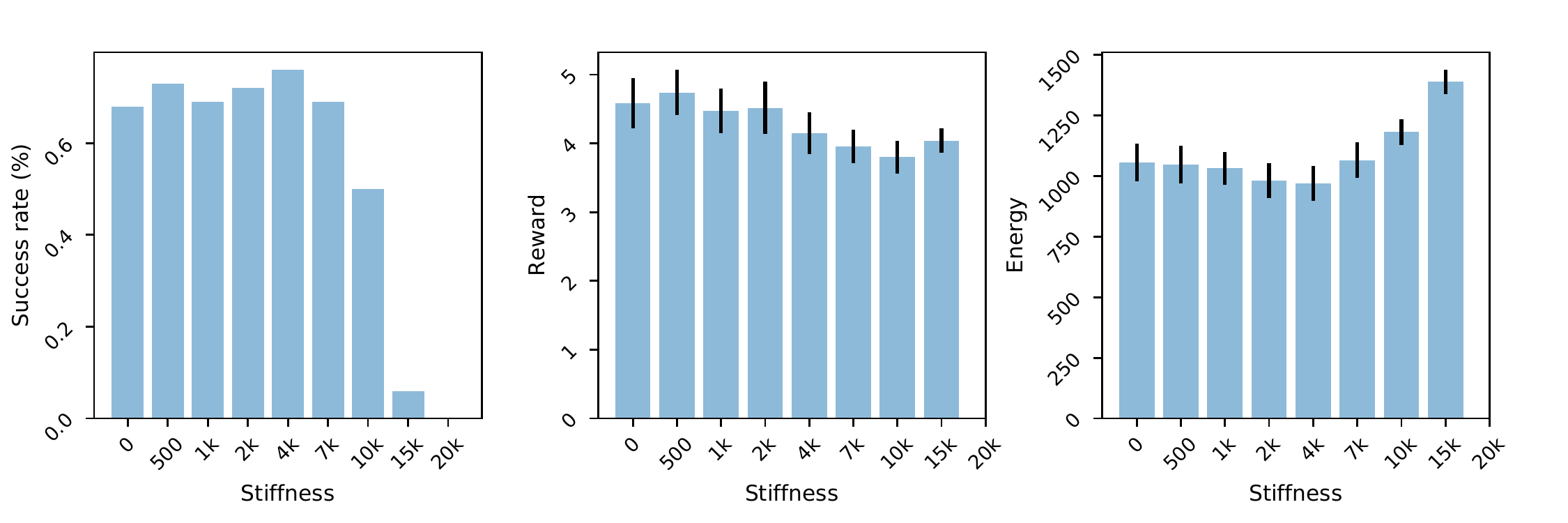}
\caption{Results of the locomotion task using the G2P algorithm.50 Monte Carlo runs for each case.}
\label{fig:Fig5}
\end{figure*}

Fig.~\ref{fig:Fig5} shows the results for the G2P implementation of the locomotion task for 50 Monte Carlo runs (also see the Supplementary Video). Fig.~\ref{fig:Fig5}a shows the success rate (if the algorithm found a solution that passes the 3m threshold within 100 exploration attempts). This figure shows that except for very high stiffness values, the algorithm could find a way to fulfil the task. Fig.~\ref{fig:Fig5}b shows the ultimate reward for the successful attempts. Since G2P algorithm is not strict on maximizing the reward (finds a good-enough solution within few attempt), we cannot see any big distinction between these final reward. Fig.~\ref{fig:Fig5}c shows the energy consumption for the attempts with the ultimate reward (here we define energy as the sum of squared activation values for all three muscles and across time). This figure shows that the energy consumption for the mid-range stiffness values is lower. It is important to note that we did not put an energy cost term in the reward and therefore, this pattern is an emergent feature of the physics of the system. This result justify future studies that would focus on utilizing stiffness in reducing energy costs.
\begin{figure*}
\includegraphics[width=1 \linewidth]{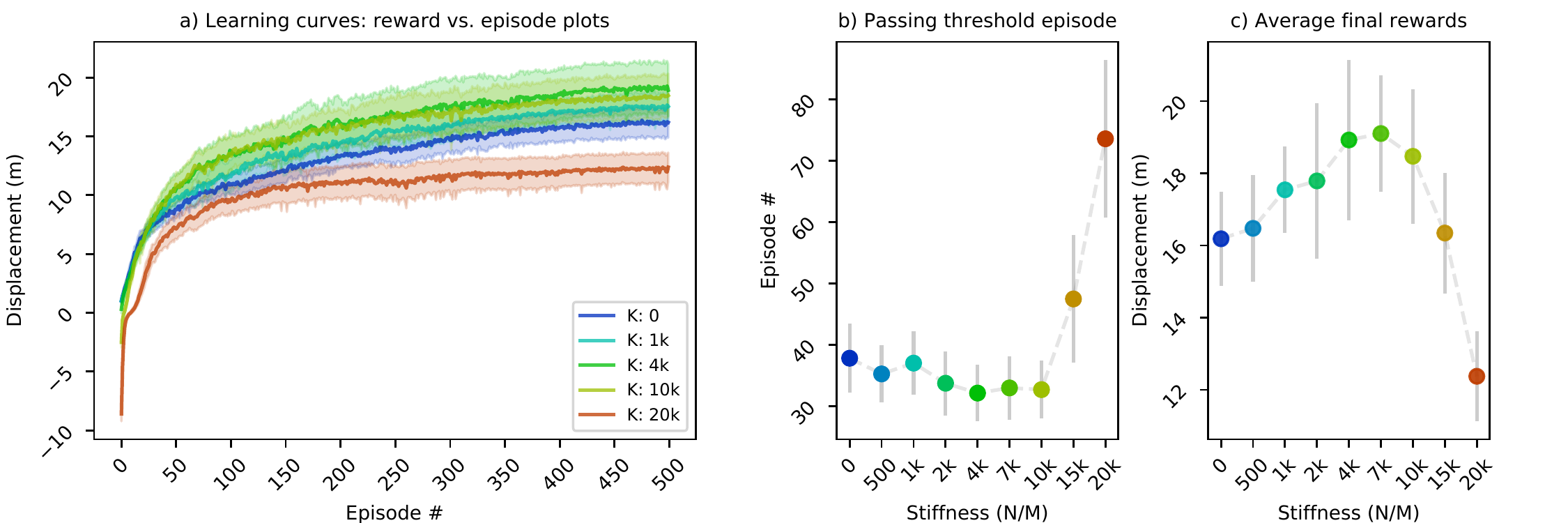}
\caption{Results of the locomotion task using the PPO algorithm, 50 Monte Carlo runs for each case.}
\label{fig:Fig6}
\end{figure*}
Lastly, Fig.~\ref{fig:Fig6} shows results for the PPO implementation of the locomotion task for 50 Monte Carlo runs (also see the Supplementary Video). Fig.~\ref{fig:Fig6}a shows that all learning curves exhibit a consistent pattern where systems with mid-range stiffness values raise faster and also end up with higher ultimate rewards. Fig.~\ref{fig:Fig6}b shows the first episode in which the Fig.~\ref{fig:Fig6}a curves passed an arbitrary reward cap (9m for this figure). The plot can slightly change based on the selected threshold but the pattern is consistent in that systems with mid-range stiffness values need less episodes to pass any reward cap. Finally, Fig.~\ref{fig:Fig6}c shows the ultimate rewards in which, again, consistent with all other findings of this paper, a mid-range stiffness value resulted in higher performance. It is important to note that although the PPO algorithm does not use an explicit inverse map, it builds an implicit inverse map which justifies why the results are consistent with the ones coming from the G2P algorithm (which uses an inverse map in an hierarchical structure).

One important point we observed in our simulations was oscillatory behaviour (chatter) in systems with very high stiffness (see Supplementary Video). The likely origin of this is that high stiffness in the muscle model, we now see, can make the system have modes at higher resonant frequencies (analogous to high gains for small errors in a proportional controller) that can lead to instability and interfere with the numerical integrator.
This happens at high stiffness values even though our MuJoCo model has mild damping and frictional losses distributed throughout the body (i.e., at joints, contact model, muscles, etc.) to make the system more stable, realistic and numerically efficient.

\section{Discussion}

Here we show, first, the feasibility of autonomous learning and adaptation in the presence of elastic elements in tendon-driven systems. And second,  we provide evidence that changes in the parallel stiffness of the actuators (i.e., muscle model) affects both learning rate and performance. Our results are useful in that  they show (i) fast learning and adaptation in systems known to be challenging to control with analytical approaches, and (ii) great promise and opportunity for the design of robotic systems where tuning the stiffness of the actuators can greatly enhance performance while leveraging the inherent passive properties of elastic elements that also grant stability and potential energy efficiency with, importantly, minimal to no degradation in learning rates.

These findings are critical for the future evolution of robot design, which to date has splintered into two main camps: `conventional' robot design with stiff bodies and actuators~\cite{yoshikawa1990foundations} vs. `soft robots' that have few to no stiff elements~\cite{verl2015soft}.
Our work here now points to a third option that can, in principle, combine the benefits of both approaches by populating the spectrum between them.
In our prior work, we have emphasized that the design space of tendon-driven systems must include both the topology of the limb (i.e., the number, type and connectivity among its elements) and the parameters of the individual elements (e.g., joints, linkages and tendons) for both robots and musculoskeletal systems~\cite{inouye2014optimizing, valero2007beyond}; we have also explored the extreme case of purely data-driven locomotion of tensegrity structures~\cite{rieffel2009automated} and limbs~\cite{marjaninejad2019autonomous}. However, that work did not explicitly explore the consequences of elasticity to learning per se.

We now argue that elasticity is an inevitable element of tendon-driven robots and biological systems (see Introduction), and thus must be systematically and explicitly considered in this current AI wave seeking to develop autonomous learning for robots, and to understand neuromuscular control in animals.
As such, our results argue for, and enable, the co-development of robot bodies and autonomous controllers that take advantage of elastic elements, which can lead to improved learning and performance---while also taking advantage of its intrinsic benefits of stability, energetic efficiency, and impact absorption. It is important to underline that the main focus of this study was not to optimize for performance or energy efficiency. Moreover, we used two of the most recent algorithm that prove to be suitable for the test case in hand but similarly, other state of the art algorithms can also be used in the future to control these systems. Our results, therefore, open the door to development efforts that recapitulate the beneficial aspects of the co-evolution of brains and bodies in vertebrates.

One particularly interesting observation from Fig. \ref{fig:Fig2} is that it was initially easier for the ANN in G2P to fit to the data when the muscles had low stiffness values. And then, after a few epochs, the fit was better with higher stiffness values. This suggests that, in principle, learning would be optimized if one were to start out with low stiffnesses that increased over time. This is paralleled by the fact that most vertebrates start their life with a more compliant anatomy which stiffens with development~\cite{stenroth2012age,o2010mechanical,gosline2002elastic}. In our prior work, we have discussed in detail how the over-determined nature of tendon-driven systems with stretch reflexes in the muscles can make them difficult to control~\cite{valero2016fundamentals}. This is because the rotation of a joint will be impeded or disrupted if even one of the muscles that crosses it fails to lengthen (via its stretch reflex). That is, multiple constraints (i.e., muscle lengthenings) must be satisfied when driven by few variables (i.e., join angles). Such over-determined systems, which have more variables than equations, have at most one solution and are solved in practice via least-squares error methods. In such methods, a solution is found by finding a set of variables that violate the constraint equations the least (in an Euclidean norm or sum-of-squares sense). This is why, in the past, we have called the elasticity of musculotendons (the combinations of muscle and tendon) as a `critical enabler' of the neural control of smooth movements~\cite{valero2016fundamentals}. The results Fig. \ref{fig:Fig2} bear this out: it is easier to learn to control tendon-driven system where low stiffnesses at the muscles provide a large error margin for muscle lengths at the expense of performance; but stiffening the system once the initial learning has taken place will improve performance. This, in a sense, is a form of morphological curriculum learning that can enable new thinking about `developmental robotics,' where changes that happen within an individual's life span improve learning and performance, echoing the work of Bongard where morphological changes within a single individual aid learning~\cite{bongard2011morphological}. This is an interesting path for future work and is distinct from `evolutionary' robotics that occurs over multiple generations of individuals.

Other future work could focus on the development of hardware/software to exploit these benefits of elastic elements, especially in tendon-driven systems. This also opens up opportunities for testing autonomous learning algorithms and assessing their performance in more sophisticated designs (such as bipeds or quadrupeds, especially in their physical implementations), and more challenging tasks and environments. It is important to note that this study worked within the abilities and limitations of MuJoCo, which implements a very particular version of a Hill-Type muscle model that does not include a tendon with the elasticity and viscosity parameters of the aponeurosis and tendon~\cite{valero2016fundamentals}. The stiffness values that we changed in the muscle model are those for the parallel elastic element to the force generating module that uses a simple approximation to the force-length and force-velocity properties of muscle~\cite{todorov2012mujoco,valero2016fundamentals}, and does not contain the natural spinal closed-loop control (i.e., afferentation) of muscles~\cite{nagamori2018cardinal}. Studying the effects of the series elastic element (which, by the way, is also the stress-strain curve of a mechanical cable in a robot) would be an interesting and necessary path to follow in the future work.




\section*{CODE AVAILABILITY}
The code and the MuJoCo models used in this study and the supplementary video can be accessed through project's Github repository at: \url{https://github.com/marjanin/tendon_stiffness}

\section*{ACKNOWLEDGMENTS}
This project was supported by NIH Grants R01-052345 and R01-050520, award MR150091 by DoD, and award W911NF1820264 by DARPA-L2M program. Also, by USC Provost Fellowship to A.M. and the Consejo Nacional de Ciencia y Tecnología (Mexico) fellowship to D.U.-M.


\bibliographystyle{IEEEtran}
\bibliography{IEEEabrv,Marjaninejad_bib}

\end{document}